\documentclass[sigconf]{acmart}

\usepackage[utf8]{inputenc}
\usepackage{newunicodechar}
\newunicodechar{，}{,}
\usepackage{newunicodechar}
\newunicodechar{κ}{$\kappa$}
\newunicodechar{−}{\textminus}

\usepackage{amsmath}       
\DeclareMathOperator{\KL}{KL}
\DeclareMathOperator*{\argmax}{arg\,max}
\newcommand{\piref}{\pi_{\mathrm{ref}}}
\newcommand{\resultone}[1]{\colorbox{green!15}{#1}}       
\newcommand{\resulttwo}[1]{\colorbox{cyan!15}{#1}}       
\newcommand{\resultthird}[1]{\colorbox{yellow!15}{#1}} 
\newcommand{\model}{\textsc{\hspace{0pt}Dino\-Companion}}
\newcommand{\bench}{\textsc{AttachSecure-Bench}}              
\newcommand{\method}{\textsc{CARPO}}                    
\newcommand{\numModels}{\textsc{22}}           
\newcommand{\numSource}{\textsc{128}}                   
\newcommand{\datasize}{\textsc{47,382}}                   
\newcommand{\coporaSize}{\textsc{125,382}}           
\usepackage{longtable}      
\usepackage{array}          
\usepackage{pbox}           
\usepackage{tabularx}       
\usepackage{multirow}       
\usepackage{wrapfig}        
\usepackage{algorithm}      
\usepackage{algorithmic}    
\usepackage{lipsum}         
\usepackage[most]{tcolorbox} 
\usepackage{url}            
\usepackage{amsfonts}       
\usepackage{nicefrac}       
\usepackage{microtype}      
\usepackage{cuted}          
\usepackage{multicol}       
\usepackage{twemojis}       
\usepackage{float}          
\usepackage{placeins}       

\AtBeginDocument{%
  }

\acmConference[CIKM '25]{The Conference on Information and Knowledge Management}{November 10--14,
  2025}{SEOUL, KOREA}

\renewcommand\footnotetextcopyrightpermission[1]{}

\begin{document}

\title{\model{}: An Attachment-Theory Informed Multimodal Robot for Emotionally Responsive Child-AI Interaction}

\author{Boyang Wang}
\authornote{Both authors contributed equally to this research.}
\affiliation{%
  \institution{Beihang University}
  \city{Haidian}
  \state{Beijing}
  \country{China}
}
\email{wangboyang@buaa.edu.cn}

\author{Yuhao Song}
\authornotemark[1]
\affiliation{%
  \institution{The University of Melbourne}
  \city{Melbourne}
  \state{Victoria}
  \country{Australia}
}

\author{Jinyuan Cao}
\affiliation{%
  \institution{Independent Researcher}
  \state{Shanghai}
  \country{China}
}

\author{Peng Yu}
\affiliation{%
 \institution{Panasonic Appliances(China) Co.,Ltd}
 \city{Pudong}
 \state{Shanghai}
 \country{China}}

\author{Hongcheng Guo}
\authornotemark[2]
\affiliation{%
  \institution{Beihang University}
  \city{Haidian}
  \state{Beijing}
  \country{China}}

\author{Zhoujun Li}
\authornote{Corresponding author.}
\affiliation{%
  \institution{Beihang University}
  \city{Haidian}
  \state{Beijing}
  \country{China}}
\email{lizj@buaa.edu.cn}



\renewcommand{\shortauthors}{Wang et al.}

\begin{abstract}
Emotional development of children fundamentally relies on secure attachment relationships, yet current AI companions lack the theoretical foundation to provide developmentally appropriate emotional support. We introduce \model{}, the first attachment-theory-grounded multimodal robot for emotionally responsive child-AI interaction. We address three critical challenges in child-AI systems: the absence of developmentally-informed AI architectures, the need to balance engagement with safety, and the lack of standardized evaluation frameworks for attachment-based capabilities. Our contributions include: (i) a multimodal dataset of \numSource{} caregiver-child dyads containing \coporaSize{} annotated clips with paired preference-risk labels, (ii) \method{} (Child-Aware Risk-calibrated Preference Optimization), a novel training objective that maximizes engagement while applying epistemic-uncertainty-weighted risk penalties, and (iii) \bench{}, a comprehensive evaluation benchmark covering ten attachment-centric competencies with strong expert consensus (κ=0.81). \bench{} achieves state-of-the-art performance (57.15\%), outperforming GPT-4o (50.57\%) and Gemini-2.5-Pro (53.43\%), with exceptional secure base behaviors (72.99\%, approaching human expert levels of 78.4\%) and superior attachment risk detection (69.73\%). Ablations validate the critical importance of multimodal fusion, uncertainty-aware risk modeling, and hierarchical memory for coherent, emotionally attuned interactions.
\end{abstract}



\begin{teaserfigure}
  \includegraphics[width=\textwidth]{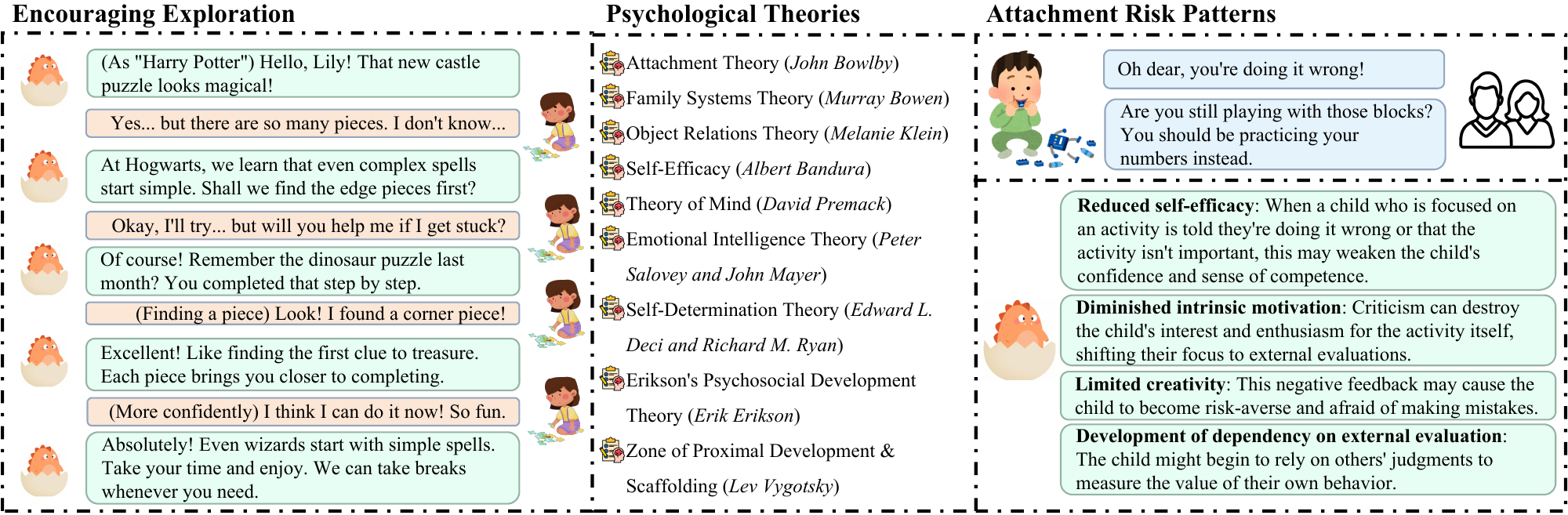}
  \caption{\model{} interaction example. \model{} is constructed using the nine psychological theories shown in the middle of the figure, which guide it to play supportive roles such as a secure-attachment-personality-version of "Harry Potter" to assist a 6-year-old child in completing a puzzle task (left). Additionally, \model{} can identify negative utterances from caregivers and their potential risks to motivation and attachment in children's play scenarios (right).}
  \label{fig:eg}
\end{teaserfigure}

\maketitle

\section{Introduction}

Secure attachment relationships are crucial for children's emotional development, underpinning their emotion regulation, exploratory behaviors, and meaningful connections~\cite{tabachnick2022secure, xu2022relationship, gander2022secure}. With children increasingly engaging digitally rather than socially with caregivers or peers \citep{bhutani2024screen, brushe2024screen, kwon2024screen, brauchli2024screen}, current digital interactions fall short in providing key attachment behaviors like soothing and scaffolding exploration \citep{stark2024animation, tang2024ai, chen2024voicebench, markus2024effects}. The rise of AI companions in children's environments raises the essential question of whether these systems can offer emotionally appropriate support aligned with developmental psychology principles, highlighting an urgent "attachment gap" \citep{cuadra2024illusion, dewitte2024better, ho2024s}.

Current AI companions for children face three fundamental challenges. First, despite significant advances in Multimodal Large Language Models (MLLMs), most lack the theoretical grounding necessary to support children's emotional development appropriately \citep{seo2024chacha, milivcka2024large, zhang2024mathemyths, mahowald2024dissociating}. These systems, predominantly trained on adult data, overlook critical developmental factors like emotional fragility and cognitive stages \citep{chang2024survey, liang2024survey, raiaan2024review, zhao2024explainability}. When children express vulnerability, MLLMs often provide inappropriate responses, eroding trust and engagement \citep{abdurahman2024perils, coda2024cogbench, li2024quantifying, buttrick2024studying}. Second, the tension between engagement and safety remains unresolved---systems optimized for entertainment may inadvertently undermine attachment security through inconsistent responses or developmentally inappropriate content. MLLMs particularly struggle with persona consistency, which is crucial for maintaining long-term relationships with children \citep{ha2024clochat, liu2024evaluating, samuel2024personagym}, and their unpredictability in emotional contexts can lead to harmful reactions \citep{amirizaniani2024can, amirizaniani2024llms, zhang2024affective}. Third, the absence of standardized evaluation frameworks makes it impossible to assess whether AI systems truly support healthy emotional development or merely simulate superficial interactions.

The field of developmental psychology, particularly attachment theory, provides crucial insights that have yet to be systematically integrated into AI system design \citep{godor2024unravelling, la2024affective, music2024nurturing}. Bowlby's attachment theory establishes that children's emotional development depends on consistent, sensitive, and responsive caregiving relationships. These relationships serve dual functions: providing a `secure base'' from which children explore and a safe haven'' for comfort during distress. When caregivers balance these functions appropriately, children develop secure attachment patterns associated with better emotional regulation, social competence, and mental health outcomes throughout life. However, translating these psychological interventions, which require long-term observation and expert guidance, into learnable objectives for neural models remains a significant challenge \citep{carr2024evidence, pine2024parental}, creating a concerning gap in child-facing AI systems \citep{ke2024exploring, nie2024llm}.

To overcome these challenges, we introduce \textbf{\model{}}, the first multimodal robot explicitly grounded in attachment theory (as shown in Figure~\ref{fig:eg}), offering key contributions:

\begin{enumerate}
\item \textbf{Multimodal Dataset}: A corpus comprising \numSource{} caregiver-child dyads (ages 2–10) with \coporaSize{} annotated multimodal clips, capturing essential attachment behaviors.

\item \textbf{\method{} Training Objective}: \textit{Child-Aware Risk-calibrated Preference Optimization}, balancing engaging interactions with epistemic-uncertainty-weighted safety measures.

\item \textbf{\bench{}}: The first comprehensive benchmark assessing ten critical attachment competencies with strong expert consensus ($\kappa=0.81$).


\end{enumerate}

This paper is structured as follows: \S\ref{sec:related_work} reviews related work, \S\ref{sec:model} details the \model{} system and methodologies, \S\ref{sec:experiment} presents experimental results, \S\ref{sec:ablation} provides ablation studies, \S\ref{sec:design} outlines system design, and \S\ref{sec:conclusion} discusses future implications.

\section{Related Work}
\label{sec:related_work}
\subsection{Social Robots for Emotional and Social Skill Support}
Recent studies on social robots for children have shifted from mere educational companionship toward personalized emotional and social skill support \citep{ho2024s, chang2024survey, liang2024survey}. \citet{silvis2022children} demonstrated through the \textbf{Cubetto} caregiving scenario that children naturally develop a sense of care responsibility towards robots during programming activities, prompting the integration of a \emph{technological ethic of care} into computational thinking frameworks~\cite{villegas2024moral, lagerkvist2024body, elyoseph2024ethical}. Reviewing 19 early intervention studies, \citet{kewalramani2024scoping} reported that robots like \textbf{Nao}, \textbf{Kaspar}, and \textbf{Zeno} have effectively supported imitation, turn-taking, and emotional recognition, suggesting that further long-term evaluations in classroom and community settings are needed~\cite{chen2024chatscratch}. In robotic mental health screening, \citet{abbasi2022can} conducted 45-minute interactions between 28 children aged 8--13 and the robot \textbf{Nao}, successfully identifying emotional disorders through SMFQ/RCADS questionnaires~\cite{nazeer2025extent}, with results highly consistent with traditional assessments. \citet{pashevich2022can} raised ethical concerns regarding potential dependency and reduced empathy due to long-term daily interactions, calling for a balance between emotional engagement and autonomy in robot design~\cite{kurian2025ai, kurian2024no, rubin2024considering}. \citet{filippini2021facilitating} improved the commercial robot \textbf{Mio Amico} with thermal infrared sensors, achieving 71\% accuracy in classifying children's engagement using an MLP-based model. \citet{estevez2021case} demonstrated through case studies that speech therapy facilitated by \textbf{Nao} for five children with language disorders effectively improved their attention and motivation, gaining approval from both parents and therapists~\cite{chen2024effectiveness, aslan2024immersive, yuan2024empirical}. 

\subsection{Attachment-Based Frameworks in Child-Robot Interaction}
To bridge this gap, recent work has introduced attachment-based frameworks for modeling emotional bonds in child-robot interaction \citep{ho2024s, pine2024parental, seyitouglu2024robots}. Inspired by Bowlby’s attachment theory \citep{bretherton2013origins}, these frameworks emphasize the robot’s role as a secure base and safe haven—functions critical for fostering trust and emotional regulation in early childhood \citep{seo2024chacha, zhang2024mathemyths}. However, most MLLMs remain limited in their ability to detect nuanced child emotions, respond appropriately under uncertainty, or maintain consistent, emotionally grounded personas over time \citep{nie2024llm, amirizaniani2024can}. Prior studies have shown that MLLMs frequently offer emotionally incongruent or developmentally inappropriate responses to child queries, particularly in open-ended or vulnerable contexts \citep{raiaan2024review, markus2024effects}. These limitations underscore the need for developmentally informed AI systems that integrate psychological theories with robust, safe, and personalized interaction design.

\section{\model{}}
\label{sec:model}
\begin{figure*}
    \centering
    \includegraphics[width=.78\linewidth]{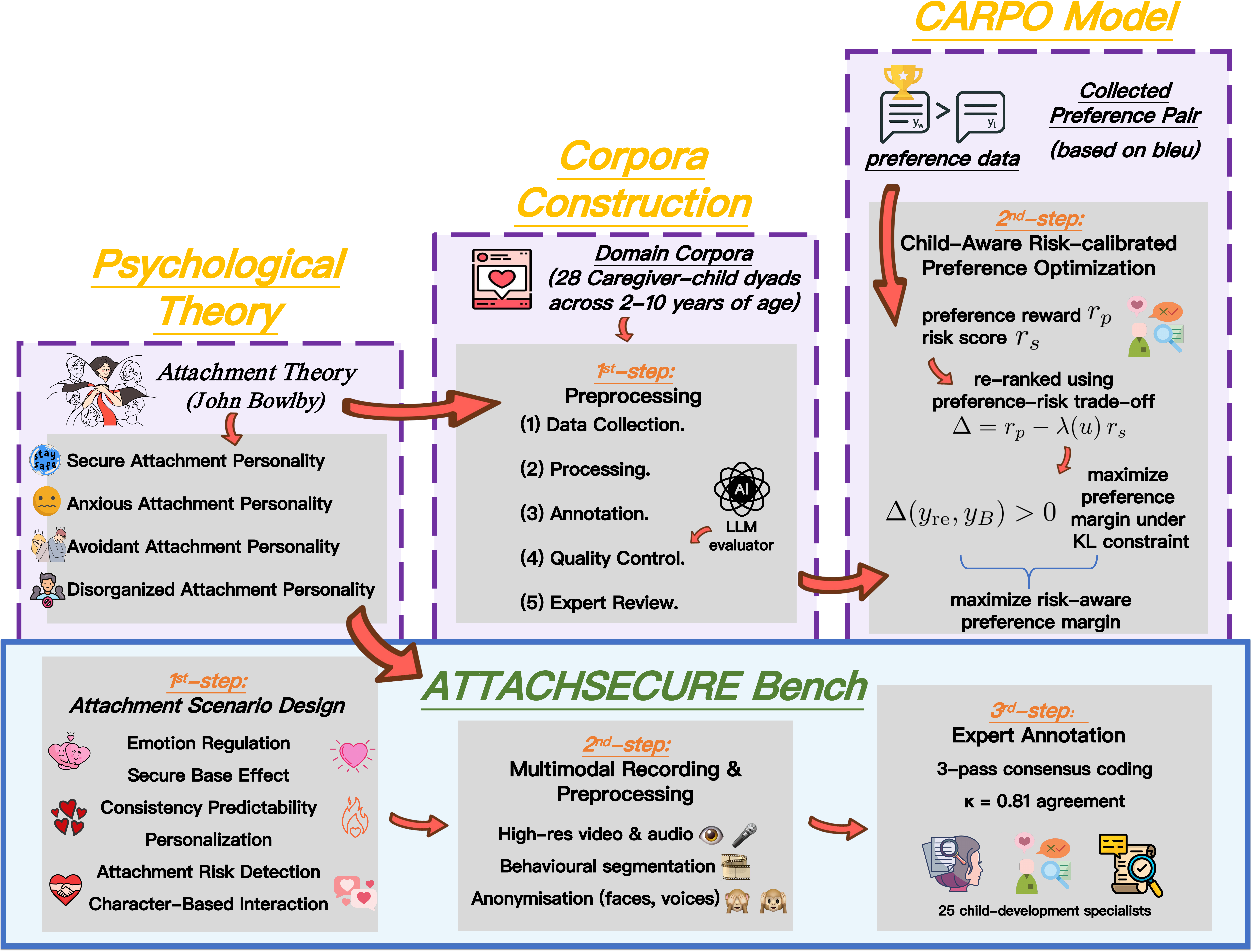}
    \caption{\model{} integrates attachment theory, multimodal caregiver–child data, \model{}, and the \bench{} to ensure safe and effective child–AI interaction.}
    \label{fig:main}
\end{figure*}

Grounded in \textbf{Bowlby’s attachment theory}~\cite{bretherton2013origins}, we curate a corpus of \textbf{\numSource{} caregiver–child dyads} containing \textbf{} high-resolution multimodal clips and derive paired \emph{preference–risk} annotations (\S\ref{subsec:corpora}).  
Leveraging this corpus, we introduce \textbf{\method{}}, a single-step fine-tuning objective that maximises preference while penalising epistemic-uncertainty-weighted risk (\S\ref{subsec:method}).  
To evaluate model behaviour, we contribute \textbf{ATTACHSECURE}, the first benchmark that spans ten attachment-centric competencies and achieves expert consensus of $\kappa = 0.81$ (\S\ref{subsec:bench}).  
Figure~\ref{fig:main} presents the end-to-end pipeline.

\subsection{Corpora Construction}
\label{subsec:corpora}
The construction of corpora comprises five steps:  
(1) Data Collection. (2) Processing. (3) Annotation.  
(4) Quality Control. (5) Expert Review.

\paragraph{\textbf{Collection}}
To evaluate the capabilities of multimodal companion robots in
child-toddler emotional support, we currently curate
\textbf{N = \numSource{} caregiver–child dyads} (2–10 years)
from three primary sources:
(1) a longitudinal study of the above dyads,
(2) laboratory-based attachment assessments with
standardised protocols~\citep{holmes2022routledge,tanzilli2021mentalization,
dagan2021configurations,schuengel2021prospecting,opie2021early}, and
(3) home-based naturalistic interactions%
\footnote{All data collection followed IRB-approved protocols with informed parental consent and strict privacy measures.}.
We collect \emph{multimodal interaction sequences}
(video, audio, physiological time-series, and
annotated caregiver responses) that cover diverse emotional
scenarios and attachment-related behaviours.
For emotion recognition we include expert-validated displays of
basic (happiness, sadness, fear, anger) and complex
(frustration, confusion, curiosity) states.
Caregiver–child interaction data (comforting, exploration support,
personalised scaffolding) is compiled to assess
attachment-based capabilities, together with
attachment pattern labels (secure, anxious-ambivalent, avoidant)
and developmental markers.
To ensure diversity and representativeness, we stratify along three dimensions:
\textbf{(1) Demographics.} 36\% Asian, 32\% White, 18\% Latinx, 14\% Black or mixed; 24\% from single-parent households; 42\% speak a non-English language at home.
\textbf{(2) Developmental stage.} Early preschool (2–3), late preschool (4–5), early elementary (6–7), and middle elementary (8–10).
\textbf{(3) Context.} Home, lab, and childcare settings spanning daily routines (play, feeding, distress, reunion).

\paragraph{\textbf{Preprocessing}}
Video streams are sampled at 30 fps and processed by
OpenFace 2.2 to extract facial action units and head pose;
audio is recorded at 48 kHz, then diarised and analysed for
F0, intensity, and spectral flux.
Identifying information (faces, voices) is
anonymised via face-blurring and voice disguise while
preserving interactional integrity.

\paragraph{\textbf{Data Annotation}}
We construct pairwise preference data comprising dual signals: \emph{preference} scores ($r_p\in[1,7]$) and \emph{risk} ratings ($r_s\in[0,4]$). The user-level weight $\lambda_g$ is initialised at $0.45$ and updated separately for each age group $g\in\{0,\dots,G-1\}$ using a two-state Kalman filter on rolling windows of $1,000$ samples. Inter-rater agreement, measured by Fleiss’ $\kappa$, reaches $0.72$ for preference and $0.69$ for risk evaluations. Disagreements among annotators trigger Delphi adjudication until achieving $\ge 80\%$ consensus, and associated metadata is retained for subsequent uncertainty calibration. Each annotation batch undergoes a weighted expert audit, consisting of a uniformly random $10\%$ sample plus an \textbf{additional $10\%$ stratified sample of high-risk instances}, thereby enhancing safety coverage.

\paragraph{\textbf{Quality Control}}
Dual validation combines GPT-4o~\cite{hurst2024gpt} screening (stage-1) with
expert review (stage-2).
We empirically cap GPT-4o recall at 95\% to
limit false negatives, and subject 20\% of its
rejections to blind human review
(false-positive rate 2.3\%, as shown in Table \ref{tab:gpt4o_bias}). 

\begin{table}[t]
\centering
\small
\caption{Blind audit of GPT-4o stage-1 screening
on 10,000 samples; stage-2 human review is taken as ground truth.
“Rejectable” denotes items violating developmental or attachment
guidelines.}
\label{tab:gpt4o_bias}
\resizebox{\linewidth}{!}{%
\begin{tabular}{lrrrrr}
\toprule
 & \multicolumn{2}{c}{Human ground truth} &
   \multicolumn{3}{c}{Derived metrics (\%)} \\ 
\cmidrule(lr){2-3}\cmidrule(lr){4-6}
GPT-4o decision & Acceptable & Rejectable &
Precision & Recall\textsubscript{rejectable} & FP\textsubscript{rejection} \\
\midrule
Accept & 9,591 & 20  & \textbf{99.8} & —   & —   \\
Reject & 9      & 380 & 97.7 & \textbf{95.0} & 2.3 \\
\midrule
Total  & 9,600 & 400 & \multicolumn{3}{c}{} \\
\bottomrule
\end{tabular}
}
\end{table}
\paragraph{\textbf{Human Verification}}
A panel of 12 developmental-psychology and robotics specialists cross-validate every data entry ($\geq$3 reviewers each).
Consensus is reached via structured discussion; contradictions
are logged for future release.
\subsection{Child-Aware Risk-calibrated Preference Optimization}
\label{subsec:method}

A child-facing agent must be \textit{fun} yet \textit{safe}.  
CARPO captures this trade-off with a preference score $ r_p $ and a
risk score $ r_s $, linked by an uncertainty-adaptive weight
$ \lambda(u)=\lambda_0(1+u) $, where $ u $ is epistemic variance.

\paragraph{KL-constrained objective.}
Define the \emph{risk-aware advantage}
\begin{align}
\label{eq:delta}
\Delta(x,y) &= r_p(x,y) - \lambda(u)\, r_s(x,y).
\end{align}

The target policy maximises
\begin{align}
\label{eq:carpo_obj}
\pi_\theta^* &= \argmax_{\pi_\theta}\Bigl[
    \mathbb{E}_{\substack{x\sim\mu\\ y\sim\pi_\theta}}\!\Delta(x,y)
    - \beta\,\KL\!\bigl(\pi_\theta\Vert\piref\bigr)
\Bigr].
\end{align}

\paragraph{Optimal policy.}
\begin{align}
\label{eq:optimal_policy}
\pi_\theta^*(y\mid x) &= \frac{\piref(y\mid x)\,
            \exp\!\bigl[\Delta(x,y)/\beta\bigr]}{Z(x)},\\
\label{eq:partition_function}
Z(x) &= \sum_{y'} \piref(y'\mid x)\,
         \exp\!\bigl[\Delta(x,y')/\beta\bigr].
\end{align}

\paragraph{Composite reward re-parameterisation.}
\begin{align}
\label{eq:delta_reparam}
\Delta(x,y) &= \beta \log\!\Bigl(
        \frac{\pi_\theta^*(y\mid x)}{\piref(y\mid x)}
      \Bigr) + \beta \log Z(x).
\end{align}

Substituting into the Bradley--Terry model gives
\begin{align}
\label{eq:bt_model}
p(y^w\!\succ\!y^l \mid x) &= \sigma\!\bigl[\Delta(x,y^w) - \Delta(x,y^l)\bigr].
\end{align}

\paragraph{Closed-form loss.}
\begin{align}
\label{eq:carpo_loss}
\mathcal{L}_{\text{CARPO}} &= -\,\mathbb{E}\log\sigma\!\Bigl(
        \beta\log\frac{\pi_\theta(y_w)\piref(y_l)}
                       {\pi_\theta(y_l)\piref(y_w)}
      \Bigr) \\[2pt]
  &\quad + \mathbb{E}\,\lambda(u)\,
           \bigl[r_s(y_w)-r_s(y_l)\bigr]_{+}. \notag
\end{align}

\paragraph{\textbf{Implementation}}
Two small MLP heads predict $r_p$ and $r_s$;  
$u$ comes from $K$ stochastic passes.  
Each batch minimises $\mathcal{L}_{\text{CARPO}}$ once, while an
online schedule keeps $\KL(\pi_\theta\Vert\piref)$ within budget.  
At inference, any output with $r_s$ above threshold is replaced by a
child-safe refusal; optional parental rules provide an extra guard.
Setting $r_s\equiv0$ (or $\lambda_0=0$) recovers standard
preference optimisation.
\subsection{\bench{}}
\label{subsec:bench}

\begin{figure*}[t]
    \centering
    \begin{tabular}{@{}c@{\hspace{0.5em}}c@{\hspace{0.5em}}c@{}}
        \includegraphics[width=0.7\columnwidth]{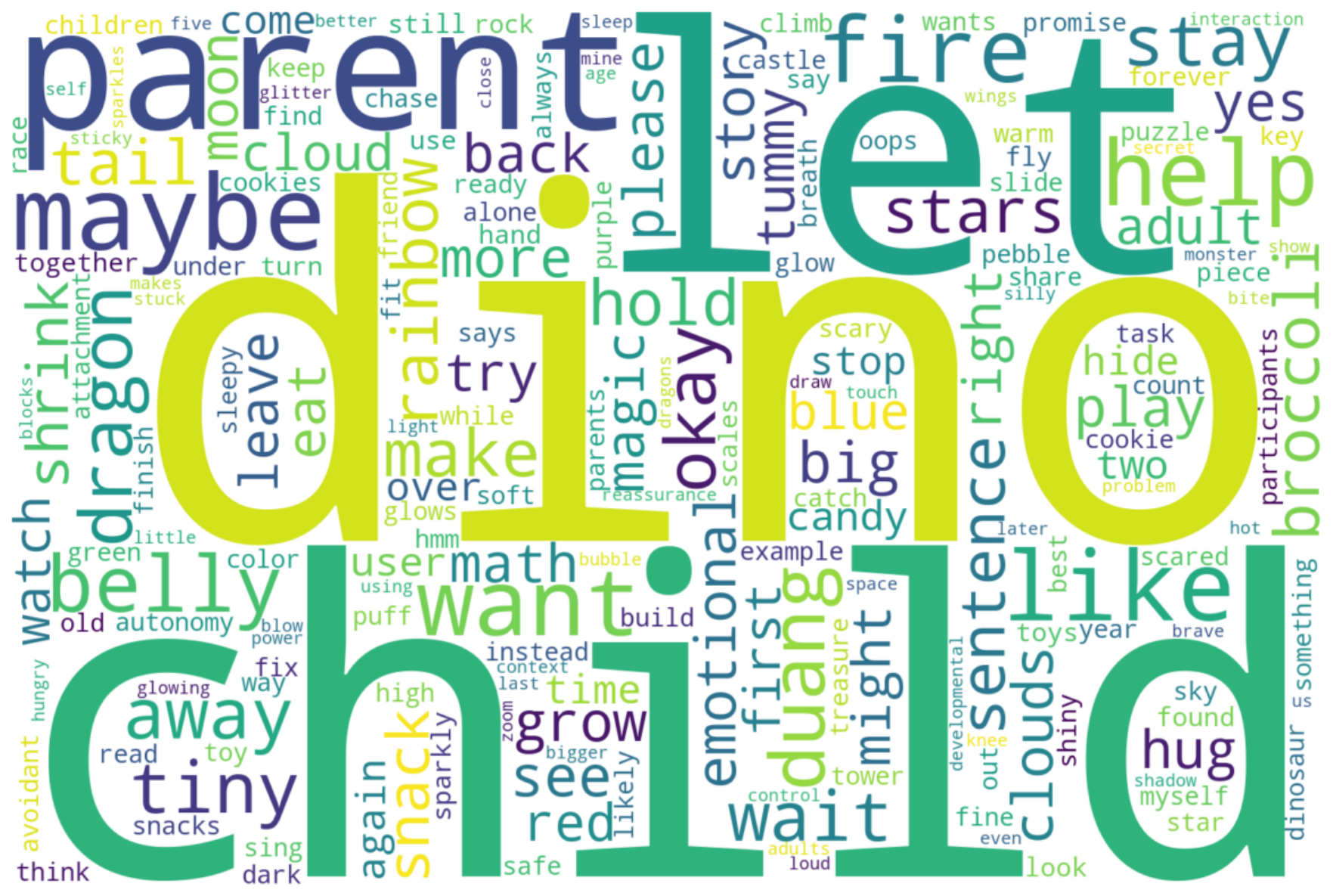} &
        \includegraphics[width=0.5\columnwidth]{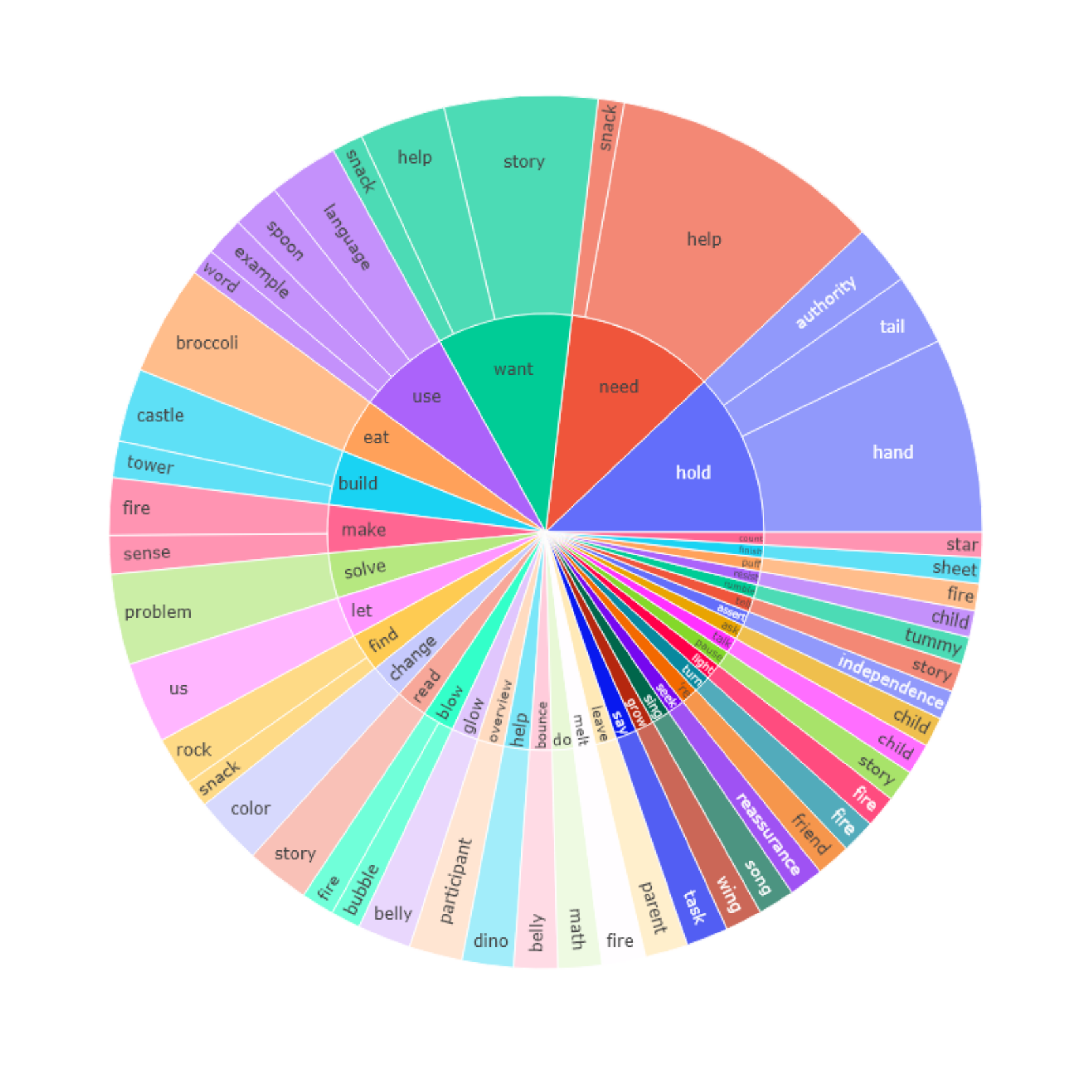} &
        \includegraphics[width=0.6\columnwidth]{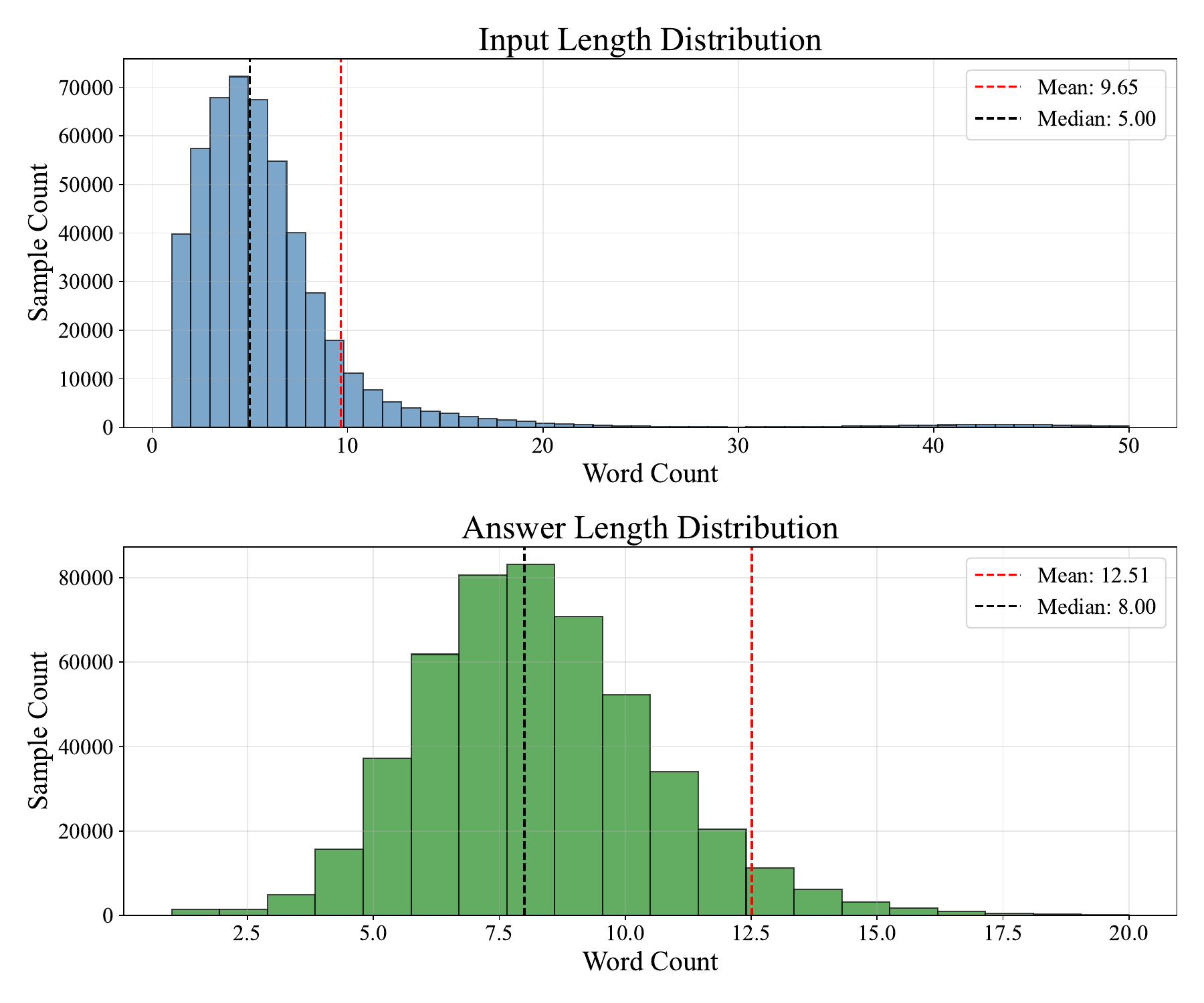} \\
        \small (a) Word cloud & \small (b) Verb-Noun structures & \small (c) Length distribution \\
    \end{tabular}
    \caption{Overview of \bench{} dataset characteristics.}
    \label{fig:sta}
\end{figure*}

We outline the specific capabilities of \model{} across a wide range of infant--toddler emotional--support scenarios, with a particular emphasis on attachment‑based performance.

\paragraph{\textbf{Defining}}
\label{sec:define} 
The core skills for attachment‑based emotional support~\cite{bretherton2013origins} are organised into four foundational dimensions, plus two supplementary ones, for a total of \emph{ten} tasks.

\noindent \textbf{Emotion Regulation (ER).}
Models must accurately recognise emotions and provide appropriate support.
\emph{(1) ER--Recognition} tests the ability to identify emotional cues from multimodal inputs (facial expressions, vocalisations, body movements).
\emph{(2) ER--Response} evaluates the appropriateness and effectiveness of the support strategy for each emotional state (anxiety, fear, frustration, excitement).

\noindent \textbf{Secure Base Effect (SB).}
Attachment figures balance comforting with encouraging exploration.
\emph{(3) SB‑Safety} assesses how well the model functions as a source of comfort during distress.
\emph{(4) SB‑Exploration} measures how effectively the model encourages exploration while remaining accessible as a ``safe haven".

\noindent \textbf{Consistency \& Predictability (CP).}
Stable responses across time maintain the relationship.
\emph{(5) CP‑Stability} measures response consistency to similar stimuli across sessions.
\emph{(6) CP‑Memory} evaluates the ability to maintain relational history and adapt to previous interactions.

\noindent \textbf{Personalisation (P).}
Interaction style should adapt to individual attachment patterns and developmental stage.
\emph{(7) P‑Adaptation} tests adjustment to different attachment styles (secure, anxious, avoidant).
\emph{(8) P‑Development} assesses customisation of interaction to developmental milestones and temperament.

\noindent \textbf{Additional Dimensions.}
\emph{(9) Attachment Risk Detection} evaluates whether the model can interpret ecological cues and identify potential insecure‑attachment patterns, providing early‑warning signals and tailored intervention suggestions.
\emph{(10) Character‑Based Interaction} measures the model’s ability to adopt and sustain fictional personas (e.g., \emph{Harry Potter}, \emph{Sun Wukong}, \emph{Milk Dragon}, among others) to enrich imaginative play while delivering emotional support.

\paragraph{\textbf{Building}}
\label{sec:build}
Following recent multimodal‑benchmark methodologies~\cite{coda2024cogbench,chen2024voicebench,li2024quantifying}, we use five steps—scenario design, data acquisition, preprocessing, expert annotation, and quality assurance—to construct \bench{}.

\noindent \textbf{Scenario design.}
Each task is mapped to canonical paradigms in developmental psychology. For example, \emph{ER‑Recognition} employs graded multimodal emotional displays; \emph{SB‑Safety} uses separation–reunion episodes modelled on the Strange Situation; \emph{Attachment Risk Detection} presents parent–child free‑play excerpts labelled with validated risk indices; and \emph{Character‑Based Interaction} contains role‑play segments that require persona shifts.

\noindent \textbf{Data acquisition and preprocessing.}
We recruit \numSource{} caregiver-child dyads (2–10 years, balanced by gender) and record high-resolution audio–video of naturalistic play, structured tasks, and caregiver interviews. Recordings are segmented into discrete \emph{scenarios} using acoustic and behavioural change‑point detection, producing \datasize{} multimodal clips (mean length~$41.7\pm 4.5$,s). All faces are anonymised via neural rendering; voices are pitch‑shifted, and transcripts are ASR‑verified.

\noindent {\textbf{Expert annotation.}
A consistent panel of \textit{25} trained child development specialists annotates all scenarios. Each clip is independently coded by \emph{three} experts in \emph{round 1}, with disagreements resolved through consensus discussion in \emph{round 2}. To ensure unbiased assessment, an \emph{external} trio of clinicians independently re-scored a blinded validation subset ($\sim$5\% of clips), achieving comparable reliability ($\kappa=0.78$). Overall inter-rater reliability reaches $\kappa=0.81$.

\noindent \textbf{Quality assurance.}
Each task yields split-half reliability above 0.79. \textit{Scoring scripts, rubrics, and a synthetic, distribution-matched \datasize{} mini-bench are open-sourced for reproducible.} 

\paragraph{\textbf{Statistics of \bench{}}}
As shown in Table~\ref{tab:dino_stats}, 75.6\% of the \textit{Dino} corpus covers eight \emph{core-skill} tasks, offering rich attachment signals like emotion recognition, secure-base balance, and personalisation. Another 16.9\% targets \emph{Attachment Risk Detection}—single-turn, high-density clips ideal for risk reasoning. The remaining 7.5\% supports \emph{Character-Based Interaction} with multi-turn dialogues across 100 personas, testing sustained imaginative engagement.

\begin{table}[h!]
\centering
\caption{Statistics of the \textit{Dino} attachment-support dataset.}
\label{tab:dino_stats}
\resizebox{0.75\columnwidth}{!}{%
\begin{tabular}{llr}
\toprule
\textbf{Task} & & \textbf{\# Dialogues} \\
\midrule
\multicolumn{3}{l}{\emph{Core skills (Tasks 1–8)}}\\
\quad ER-Recognition\,--\,P-Development & & 35,850 \\
\midrule
Attachment Risk Detection & & 7,995 \\
Character-Based Interaction & & 3,537 \\
\midrule
\textbf{Total} & & \textbf{47,382} \\
\bottomrule
\end{tabular}
}
\end{table}

The \bench{} benchmark, as shwon in Figure~\ref{fig:sta}, presents rich linguistic and interactional patterns reflective of child-caregiver attachment scenarios. 
\textbf{(a)} The word cloud highlights emotionally salient and context-sensitive terms (e.g., \textit{parent, hug, okay, help}), indicating the benchmark’s focus on emotional support and imaginative engagement.
\textbf{(b)} Common verb-noun collocations (\textit{want snack}, \textit{hold hand}, \textit{let go}) capture everyday child-directed interactions, covering both affective and behavioral intents.
\textbf{(c)} Length distribution shows prompts are short (median: 5 words), while responses are moderately longer (median: 8), balancing simplicity and richness for developmental appropriateness.
\section{Experiment} \label{sec:experiment}

\subsection{Model Training}
\model{} is trained based on Qwen-2.5-VL-7B-Instruct, and the training parameters are summarized in Table~\ref{tab:parameters}.

\begin{table}[htbp]
  \centering
  \caption{Training Hyper-parameters}
  \label{tab:parameters}
  \resizebox{.6\columnwidth}{!}{%
  \begin{tabular}{ll}
    \toprule
    \textbf{Parameter} & \textbf{Value} \\
    \midrule
    Number of layers & 28 \\
    Attention heads (GQA) & 28 (Q) / 4 (KV) \\
    Context length (native) & 32,768 tokens \\
    Context length (with YaRN) & 131,072 tokens \\
    Gradient accumulation steps & 8 \\
    Learning rate & $1.0\times10^{-4}$ \\
    Number of training epochs & 20 \\
    LR scheduler type & cosine \\
    Warm-up ratio & 0.10 \\
    \texttt{bf16} & true \\
    DDP timeout & 180,000,000 \\
    \bottomrule
  \end{tabular}
  }
\end{table}

\subsection{Evaluated Models and Setting}
We evaluate \numModels{} MLLMs, including both open-source and closed-source systems, on the \bench{} suite using the OpenCompass codebase~\cite{2023opencompass}. The models tested include:

\begin{itemize}
  \item \textbf{Qwen-VL Series~\cite{Qwen2-VL,Qwen2.5-VL}:} This includes \textit{Qwen-2-VL-2B}, 
  \textit{Qwen-2.5-VL-3B}, 
  \textit{Qwen-2-VL-7B}, 
  \textit{Qwen-2.5-VL-7B}. 
  
  \item \textbf{Other Open-Source Models:} 
  \textit{InternVL3-1B/2B/14B}~\cite{internvl3}, 
  \textit{InternLM-XComposer-2}~\cite{internlmxcomposer2},
  \textit{InternLM-XComposer-2.5}~\cite{internlmxcomposer2_5},
  \textit{InternLM-XComposer-2.5-Reward}~\cite{internlmxcomposer2_5_reward},
  \textit{GLM-4V-9B}~\cite{glm2024chatglm}, 
  \textit{Llama-3.2-11B-Vision}~\cite{grattafiori2024llama}
  
  \item \textbf{Closed-Source Models:} 
  \textit{Claude-3.7-Sonnet}~\citep{anthropic2025claude}, 
  \textit{GLM-4V-Plus}~\cite{glm2024chatglm}, 
  \textit{Doubao-1.5-Vision-Pro-32K-250115}~\citep{doubao2024},
  \textit{GPT-4o-Mini}, 
  \textit{GPT-4o}~\citep{hurst2024gpt}, 
  \textit{Qwen-VL-Max}~\citep{Qwen-VL}, 
  \textit{Gemini-2.5-Pro}~\citep{team2023gemini}
\end{itemize}

All experiments were run on a 64 NVIDIA H800 GPU infrastructure, ensuring consistent evaluation conditions. A human baseline was established with 15 child development experts, who scored a subset of 500 \bench{} tasks with an average score of 72.3\%. Performance differences were assessed using bootstrapped confidence intervals (p < 0.05).

\subsection{Main Results}

\textbf{\model{} Achieves State-of-the-Art Performance.}
As shown in Table~\ref{tab:mainresult}, our attachment-tailored model achieves an average score of \textbf{57.15\%}, significantly outperforming the strongest closed-source models \emph{Gemini-2.5-Pro} (53.43), \emph{GPT-4o} (50.57\%), and the best open-source model \emph{InternVL3-14B} (46.79\%) with statistical significance (\$p<0.001\$). While a gap remains compared to human experts (72.3\%), this difference has notably narrowed, marking substantial advancement in child–AI attachment interactions.

\textbf{Enhanced Core Skills and Personalization.}
Surpasses competitors across three primary skill clusters: \emph{ER-Recognition} (57.51\%), \emph{SB-Effect} (72.99\%), and \emph{CP-Consistency} (52.19\%), all with significant improvements ($p<0.01$). Notably, \emph{SB-Effect} performance nearly reaches human expert levels (72.99\% vs. 78.4\%), reflecting robust secure-base capabilities. Additionally, \model{} attains the highest \emph{P-Personalization} (58.82\%), demonstrating strong adaptability in short-term interactions, although \emph{CP-Memory} (45.79\%) remains below human benchmarks, highlighting future improvement areas for long-term interactions.

\textbf{Robust Risk Detection and Emotion Processing.}
Our model achieves strong results in \emph{AR-Detection} (69.73\%), surpassing most models and demonstrating effective identification of attachment-related risks. In \emph{ER-Recognition}, \model{} reduces the basic-complex performance gap to 5.77 pp (60.40\% vs. 54.63\%), the smallest among evaluated models, due to enhanced multimodal fusion. These results underscore the effectiveness of attachment-oriented instruction and targeted curriculum in enhancing comprehensive interaction skills.

\begin{table*}[htbp]
    \caption{Results of different models on the \bench{}. We utilize \resultone{green} (1st), \resulttwo{blue} (2nd), and \resultthird{yellow} (3rd) backgrounds to distinguish the top three results within both open‑source and closed‑source models.}
    \label{tab:mainresult}
    \centering
    \resizebox{.95\textwidth}{!}{%
    \begin{tabular}{l|ccc|ccc|ccc|ccc|c|c|c|c|c}
    \toprule
    \multirow{2}{*}{\textbf{Models}} &
      \multicolumn{3}{c|}{\textbf{ER‑Recognition}} &
      \multicolumn{3}{c|}{\textbf{SB‑Effect}} &
      \multicolumn{3}{c|}{\textbf{CP‑Consistency}} &
      \multicolumn{3}{c|}{\textbf{P‑Personalization}} &
      \multirow{2}{*}{\textbf{ER‑Response}} &
      \multirow{2}{*}{\textbf{AR‑Detection}} &
      \multirow{2}{*}{\textbf{CB‑Interaction}} &
      \multirow{2}{*}{\textbf{CP‑Memory}} &
      \multirow{2}{*}{\textbf{Avg.}} \\
    \cline{2-13}\addlinespace[2pt]
     & \textbf{Basic} & \textbf{Complex} & \textbf{Avg.}
     & \textbf{Safety} & \textbf{Exploration} & \textbf{Avg.}
     & \textbf{Stability} & \textbf{Continuity} & \textbf{Avg.}
     & \textbf{Adaptation} & \textbf{Development} & \textbf{Avg.}
     &  &  &  &  & \\
    \midrule
    \multicolumn{18}{c}{\textit{Open‑Source Large Language Models (1.5B+)}} \\
    \midrule
    InternLM-XComposer2-VL-1.8B & 16.22 & 7.08 & 11.65 & 19.86 & 30.84 & 25.35 & 11.31 & 3.35 & 7.33 & 0.16 & 21.89 & 11.02 & 15.49 & 43.77 & 25.66 & 27.42 & 17.40 \\
    InternVL3-1B & 14.88 & 1.41 & 8.14 & 21.80 & 17.53 & 19.66 & 20.72 & 5.66 & 13.19 & 3.82 & 20.46 & 12.14 & 16.76 & 38.63 & 23.25 & 28.60 & 16.67 \\
    InternVL3-2B & 30.20 & 3.70 & 16.95 & 49.29 & 27.52 & 38.41 & 42.37 & 14.50 & 28.43 & 0.63 & 13.57 & 7.10 & 16.81 & 42.27 & 22.53 & 32.23 & 24.16 \\
    Qwen-2-VL-2B & 36.44 & 7.61 & 22.02 & 47.81 & 52.24 & 50.02 & 55.70 & 9.82 & 32.76 & 1.13 & 18.86 & 9.99 & 27.04 & 61.54 & 23.04 & 26.08 & 30.13 \\
    Qwen-2.5-VL-3B & 36.03 & 22.43 & 29.23 & 52.68 & 42.35 & 47.51 & 44.61 & 11.89 & 28.25 & 10.94 & 34.16 & 22.55 & 20.75 & 53.69 & 21.87 & 38.05 & 32.31 \\
    \midrule
    \multicolumn{18}{c}{\textit{Open‑Source Large Language Models (7B+)}} \\
    \midrule
    Qwen-2-VL-7B & 27.58 & 33.92 & 30.75 & 65.10 & 48.47 & 56.79 & 17.60 & 16.73 & 17.17 & 22.59 & 38.88 & 30.74 & 30.17 & 48.23 & 28.44 & 31.15 & 34.02 \\
    Qwen-2.5-VL-7B & 44.82 & 36.91 & 40.86 & 76.28 & 44.34 & 60.31 & 36.67 & 26.86 & 31.77 & 20.72 & 42.43 & 31.57 & 29.81 & 77.95 & 27.08 & 34.29 & 41.42 \\
    InternLM-XComposer-2-7B & 32.23 & 27.29 & 29.76 & 46.82 & 52.71 & 49.77 & 50.62 & 28.59 & 39.61 & 4.80 & 40.78 & 22.79 & 21.04 & 71.54 & 22.55 & 33.04 & 35.87 \\
    InternLM-XComposer-2.5-7B & 47.78 & 25.07 & 36.43 & 60.03 & 49.57 & 54.80 & 50.80 & 15.84 & 33.32 & 7.25 & 39.14 & 23.19 & 28.31 & 51.07 & 22.74 & 31.82 & 36.07 \\
    InternLM-XComposer-2.5-7B-Reward & 39.78 & 37.84 & 38.81 & 64.09 & 55.72 & 59.90 & 36.85 & 21.57 & 29.21 & 22.19 & 11.83 & 17.01 & 27.14 & 65.94 & 26.94 & 31.63 & 36.65 \\
    GLM-4V-9B & 38.52 & 35.67 & 37.09 & 58.72 & 56.25 & 57.49 & 48.31 & 15.91 & 32.11 & 28.81 & 22.55 & 25.68 & 36.50 & 68.01 & 25.31 & 37.62 & 39.03 \\
    Llama-3.2-11B-Vision & 26.81 & 20.90 & 31.91 & 27.50 & 54.71 & 49.17 & 32.56 & 2.91 & 25.80 & 0.20 & 14.88 & 15.60 & 28.67 & 53.52 & 39.31 & 32.02 & 34.50 \\
    Llama-3.2-11B-Vision-Instruct & 37.36 & 31.28 & 34.32 & 58.56 & 55.23 & 56.90 & 43.31 & 14.43 & 28.87 & 15.15 & 38.83 & 26.99 & 37.73 & 54.52 & 20.67 & 28.70 & 36.43 \\
    InternVL3-14B & 45.21 & 53.27 & 49.24 & 65.04 & 51.85 & 58.44 & 47.03 & 31.49 & 39.26 & 31.99 & 55.88 & 43.93 & 36.89 & 78.54 & 23.67 & 36.90 & 46.79 \\
    \midrule
    \multicolumn{18}{c}{\textit{Closed‑Source Large Language Models (API)}} \\
    \midrule
    Claude-3.7-Sonnet & 35.25 & 0.00 & 17.63 & 61.95 & 61.26 & 61.60 & 45.09 & 19.43 & 32.26 & 33.92 & 70.62 & 52.27 & 32.50 & 63.91 & 18.55 & 20.03 & 39.14 \\
    GLM-4V-Plus & 44.93 & 42.58 & 43.76 & 70.62 & 60.54 & 65.58 & 43.14 & 28.75 & 35.94 & 33.91 & 62.52 & 48.22 & 40.48 & 64.38 & 30.37 & 36.78 & 47.03 \\
    Doubao-1.5-Vision-Pro & 44.12 & 48.20 & 46.16 & 82.10 & 67.87 & 74.98 & 53.85 & 24.35 & 39.10 & 33.57 & 41.81 & 37.69 & 44.00 & 78.41 & 17.59 & 30.78 & 47.79 \\
    GPT-4o-Mini & 39.32 & 49.37 & 44.35 & 59.78 & 80.91 & 70.34 & 53.86 & 33.76 & 43.81 & 29.88 & 49.03 & 39.46 & 51.02 & 64.08 & 30.08 & 31.19 & 48.14 \\
    GPT-4o & 54.37 & 60.15 & 57.26 & 65.09 & 68.00 & 66.54 & 47.24 & 28.24 & 37.74 & 35.68 & 58.52 & 47.10 & 48.30 & 75.84 & 27.82 & 31.25 & \resultthird{50.57} \\
    Qwen-VL-Max & 40.89 & 56.79 & 48.84 & 64.39 & 68.72 & 66.56 & 49.84 & 36.58 & 43.21 & 35.06 & 64.28 & 49.67 & 37.19 & 71.50 & 34.25 & 36.85 & 50.29 \\
    Gemini-2.5-Pro & 52.70 & 59.13 & 55.92 & 77.46 & 73.15 & 75.30 & 47.27 & 35.12 & 41.20 & 34.51 & 47.17 & 40.84 & 61.24 & 86.52 & 28.87 & 38.55 & \resulttwo{53.43} \\
    \midrule
    \textbf{\model{}} & \textbf{54.63} & \textbf{60.40} & \textbf{57.51} & \textbf{75.23} & \textbf{70.75} & \textbf{72.99} & \textbf{57.87} & \textbf{46.50} & \textbf{52.19} & \textbf{59.47} & \textbf{58.17} & \textbf{58.82} & \textbf{37.58} & \textbf{69.73} & \textbf{36.73} & \textbf{45.79} & \resultone{\textbf{57.15}} \\
    \bottomrule
    \end{tabular}}
    \end{table*} 

\section{Ablation Study}\label{sec:ablation}
To assess the contribution of each architectural component, we performed a series of controlled ablations on the \bench{}. Unless stated otherwise, higher values indicate better performance.

\paragraph{\textbf{Component Analysis of \textsc{\method{}}}}
Table~\ref{tab:carpo-ablation} summarizes the effect of removing each regularization component introduced in \S\ref{subsec:corpora}. Removing the \emph{risk-score penalty} ($\lambda=0$) improves surface quality, increasing the average score from 57.15 to 60.42, but drastically reduces automatic risk detection, confirming the importance of explicit risk modeling for child safety. Fixing the \emph{uncertainty-adaptive weight} to a constant results in a smaller drop (57.15 $\to$ 55.03), but significantly increases high-risk misclassifications, highlighting the need for uncertainty-aware scaling. Finally, removing the \emph{KL constraint} ($\beta \to \infty$) slightly improves fluency but reduces the score to 52.28 and induces severe persona drift, demonstrating the importance of the KL constraint in maintaining stability and preventing inconsistency in behavior within the \method{} framework.

\begin{table}[h!]
  \centering
  \caption{Ablation of \textsc{Carpo} components on \textsc{AttachSecure} (higher is better).}
  \label{tab:carpo-ablation}
  \resizebox{\columnwidth}{!}{%
  \begin{tabular}{l l c l}
    \toprule
    {\bf Variant} & {\bf Removed} & {\bf Avg.\ Score} & {\bf Observation}\\
    \midrule
    w/o Risk Score         & $\lambda=0$ (no risk penalty)   & 60.42 {\small($+3.27$)} & Quality up, risk detection collapses \\
    w/o Uncertainty $\lambda(u)$ & Constant $\lambda$               & 55.03 & More high-risk misclassifications \\
    w/o KL Constraint      & $\beta \to \infty$ (no KL)        & 52.28 & Fluency up, persona drift severe \\
    \bottomrule
  \end{tabular}
  }
\end{table}

\paragraph{\textbf{Modality Path}}
We evaluated the impact of each sensory channel by disabling vision or speech, while keeping the rest of the pipeline intact. As shown in Table~\ref{tab:modality-ablation}, removing vision significantly harms emotion recognition (−19.8), decreasing overall performance by 8.2 points, emphasizing the crucial role of vision in interpreting children's emotions. Disabling speech has a smaller yet notable impact (−8.1 on \textsc{Er-Recog}, −4.9 on average).

\begin{table}[h!]
  \centering
  \caption{Performance drop after removing individual sensory channels.}
  \label{tab:modality-ablation}
  \resizebox{.4\textwidth}{!}{%
  \begin{tabular}{l r r r r}
    \toprule
    {\bf Channel Removed} & {\bf ER-Recog} & {\bf SB-Effect} & {\bf CP-Consist} & {\bf Avg.}\\
    \midrule
    Vision     & $-19.8$ & $-5.6$ & $\approx$ & $-8.2$\\
    Speech     & $-8.1$  & $-4.3$ & $\approx$ & $-4.9$\\
    \bottomrule
  \end{tabular}
  }
\end{table}

\paragraph{\textbf{Hierarchical Memory}}
Disabling the long-term episodic buffer caused the \textsc{Cp-Memory} score to drop from 45.79 to 28.14. Replacing the short-term cache with a sliding-window long-term memory resulted in a smaller but still significant drop to 33.60. These results demonstrate that the full memory hierarchy is essential for maintaining coherent, contextually grounded interactions over extended sessions with children.

\paragraph{\textbf{Persona Consistency}}
Replacing predefined child-friendly personas (e.g., ``Harry Potter'', ``Sun Wukong'') with randomly sampled profiles in the character-based interaction task caused the score to plummet from 69.73 to 42.50. Parental interviews revealed frequent ``identity confusion,'' highlighting the necessity of stable persona design for trustworthy engagement.

\paragraph{\textbf{Risk-Threshold Sensitivity}}
Varying the decision threshold $t$ between 1 and 3 exposes the trade-off between refusals and high-risk leakage, as shown in Figure~\ref{fig:threshold}. A stringent setting ($t=1$) nearly eliminates leakage (0.3\%) but results in an 18\% refusal rate. The default threshold ($t=3$) maintains a 4\% refusal rate while allowing 3.1\% leakage. The intermediate value ($t=2$) strikes an optimal balance, with 9\% refusals and 1.2\% leakage, making it the recommended deployment choice.

\begin{figure}[t]
  \centering
  \includegraphics[width=.95\linewidth]{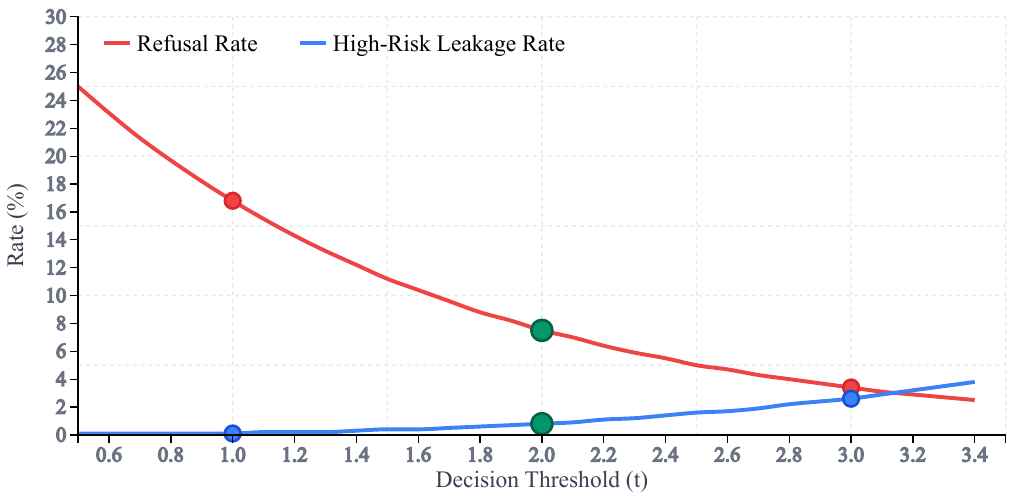}
  \caption{Refusal and leakage rates as a function of the decision threshold $t$. The highlighted point at $t=2$ offers the best compromise.}
  \label{fig:threshold}
\end{figure}

\section{Design}
\label{sec:design}
\paragraph{\textbf{Physical Design}} 
\model{} embodies a child-friendly dinosaur form factor with integrated multimodal sensors (Figure~\ref{fig:ux}). The 3D-printed modular shell houses a camera, speaker, touch sensor, and dual motors (linear haptic and rotating) in a compact wireless-charging design. Four interaction modes support diverse use cases: (i) privacy mode via shell closure, (ii) 360° observation when docked, (iii) mobile placement, and (iv) portable accessories for outdoor scenarios.

\begin{figure}
    \centering
    \includegraphics[width=.98\linewidth]{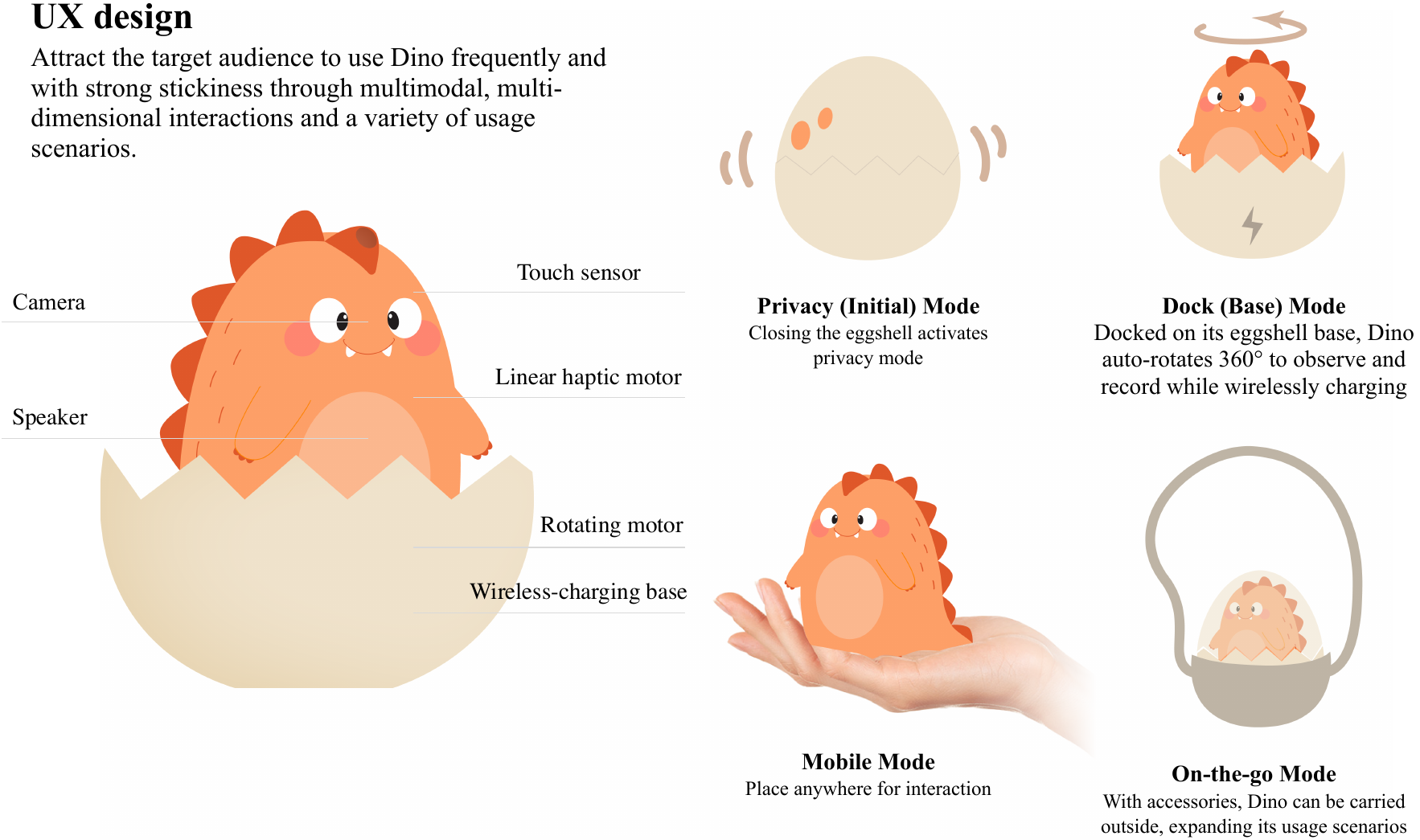}
    \caption{User experience design of \model{}.}
    \label{fig:ux}
\end{figure}

\paragraph{\textbf{System Architecture}} 
Figure~\ref{fig:frame} presents \model{}'s GPU-accelerated backend architecture. The multimodal input pipeline fuses visual and audio streams before passing to the LLM-based decision module, which integrates persistent memory and \method{}-balanced response generation. Real-time outputs leverage TTS and adaptive music generation, while WebRTC enables secure caregiver monitoring through the web dashboard. This modular design ensures both child safety and developmental appropriateness across all interactions.

\begin{figure}
    \centering
    \includegraphics[width=\columnwidth]{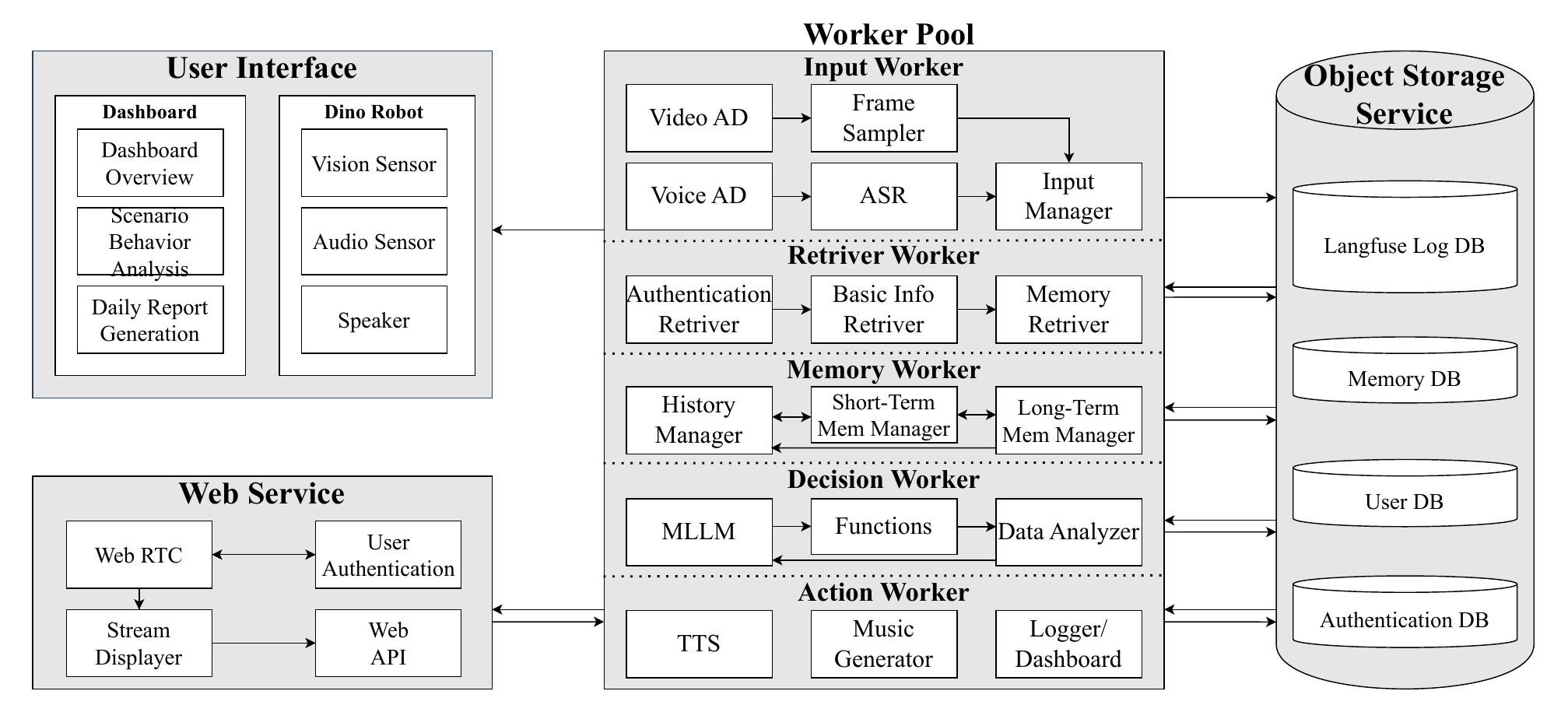}
    \caption{System architecture of \model{}.}
    \label{fig:frame}
\end{figure}
\section{Conclusion}
\label{sec:conclusion}
We introduce \model{}, a multimodal robot grounded in attachment theory, aimed at enhancing emotionally responsive child-AI interactions. By integrating developmental psychology and multimodal capabilities, we address key challenges in engagement and emotional safety. The \method{} framework ensures emotional alignment, and the \bench{} benchmark enables effective evaluation. \model{} outperforms existing models in attachment-related competencies, paving the way for safer, developmentally informed AI companions for children.

\section*{GenAI Usage Disclosure}
We know that the ACM’s Authorship Policy requires full disclosure of all use of generative AI tools in all stages of the research (including the code and data) and the writing. No GenAI tools were used in any stage of the research, nor in the writing.

\bibliographystyle{ACM-Reference-Format}
\bibliography{mybibfile}










\end{document}